\documentclass[conference]{IEEEtran}
\IEEEoverridecommandlockouts
\usepackage{cite}
\usepackage{amsmath,amssymb,amsfonts}
\usepackage{algorithmic}
\usepackage{graphicx}
\usepackage{textcomp}
\usepackage{xcolor}
\def\BibTeX{{\rm B\kern-.05em{\sc i\kern-.025em b}\kern-.08em
    T\kern-.1667em\lower.7ex\hbox{E}\kern-.125emX}}
\begin{document}

\title{Classification Methods Based on Machine Learning for the Analysis of Fetal Health Data \\
}
\author{\IEEEauthorblockN{Binod Regmi}
\IEEEauthorblockA{\textit{Department of Physics and Astronomy} \\
\textit{Mississippi state university}\\
Mississippi State, MS 39762, USA \\
regmibinod53@gmail.com}
\and
\IEEEauthorblockN{Chiranjibi Shah}
\IEEEauthorblockA{\textit{Department of Electrical and Computer Engineering} \\
\textit{Mississippi State University}\\
Mississippi State, MS 39762, USA\\
chiranjibishahmsu@gmail.com}

}

\maketitle

\begin{abstract}
The persistent battle to decrease childhood mortality serves as a commonly employed benchmark for gauging advancements in the field of medicine. 
Globally, the under-5 mortality rate stands at approximately 5 million, with a significant portion of these deaths being avoidable. Given the significance of this problem,  Machine learning-based techniques have emerged as a prominent tool for assessing fetal health. In this work, we have analyzed the classification performance of various machine learning models for fetal health analysis. Classification performance of various machine learning models, such as support vector machine (SVM), random forest(RF), and attentive interpretable tabular learning (TabNet) have been assessed on fetal health. Moreover, dimensionality reduction techniques, such as Principal component analysis (PCA) and Linear discriminant analysis (LDA) have been implemented to obtain better classification performance with less number of features. A TabNet model on a fetal health dataset provides a classification accuracy of 94.36\%. In general, this technology empowers doctors and healthcare experts to achieve precise fetal health classification and identify the most influential features in the process.
\end{abstract}

\begin{IEEEkeywords}
 Fetal health, Machine learning, principal component analysis (PCA), linear discriminant analysis (LDA), support vector machine (SVM), random forest (RF), attentive interpretable tabular learning (TabNet).
\end{IEEEkeywords}
\section{Introduction}
Premature birth (PTB) represents a significant public health concern with far-reaching implications for both individuals and communities \cite{cho22fet}. This phenomenon is characterized by its distinct nature, contributing to adverse outcomes for both families and society. Globally, neonatal mortality and morbidity rank as the primary contributors to infant fatalities and illnesses, making them the second most prevalent cause of infant mortality in developing nations. Pregnancy and childbirth have provided opportunities for medical interventions, prompting professionals and scholars to explore various successful approaches to reduce the incidence of premature births and complications among expectant mothers. Healthcare services play a crucial role in these endeavors, with preventive measures offered to all pregnant women to mitigate the risk of preterm birth and other medical issues. Interventions focus on enhancing women's awareness of early pregnancy symptoms that may indicate potential difficulties. Maternal history is a vital aspect of the examination process for pregnant women, while neonatal research investigates specific therapeutic interventions for newborns. Assessing the health, illnesses, and care provided to newborns is an integral part of this research. For several decades, infant mortality has remained a persistent concern within healthcare systems worldwide. While advancements have been made in developing tools to evaluate various aspects of fetal well-being, the interpretation of cardiotocography (CTG) data can pose challenges, particularly in regions lacking expert obstetricians \cite{yin23us}. Even in areas with access to medical professionals, the process of individually diagnosing fetuses based on CTG measurements can be time-consuming and generally inefficient. However, the application of machine learning models allows for fetal health classifications to be made without the presence of obstetricians and in a more efficient manner. These models have demonstrated high accuracy in their predictions, presenting viable solutions to the challenges surrounding fetal health.  Machine learning techniques play a crucial role in extracting valuable knowledge and uncovering hidden insights from the available system data. These techniques contribute to the development of efficient medical decision-making systems, leveraging various tools and technologies to construct algorithms for this purpose. Despite the theoretical efficacy of this approach, there have been significant hurdles in implementing machine learning models in practice.
   
To address these challenges effectively, the implementation of an explainable model is considered the most efficient approach. Such a model not only achieves accurate predictions but also provides insights into the decision-making process, enabling scientists and researchers to understand its reasoning. This knowledge equips obstetricians to communicate specific abnormal metrics to their patients, facilitating improved patient care. For instance, if the model predicts a pathological case for a fetus, it can also indicate that the prediction is based on a low frequency of uterine contractions per second. Armed with this information, a doctor can advise the patient on appropriate measures such as rest and hydration, and in severe cases, administer drugs like Oxytocin to restore normal levels \cite{gill2023abnormal}.

 To attain a model that is both high-performing and explainable, the implementation involved three distinct models: support vector machine (SVM) \cite{shah21col,gao15svm,shah22spat}, random forest(RF)\cite{shah22tab,tin98rand}, and attentive interpretable tabular learning (TabNet)\cite{shah22tab,arik20tab}. In addition, dimensionality reduction techniques, such as Principal component analysis (PCA) \cite{chris06patt} and Linear discriminant analysis (LDA) \cite{li14dec} have been implemented for obtaining better classification accuracy on fetal health with a reduced number of features.

 This paper presents the use of TabNet \cite{shah22tab,arik20tab}, a novel deep neural architecture specifically designed for tabular data, in the classification of fatal health conditions. When dealing with large datasets, employing a deep neural network (DNN) can enhance classification performance by enabling end-to-end learning through gradient descent. In contrast, tree learning methods lack the utilization of backpropagation and error signals for guiding inputs, leading to performance limitations when dealing with extensive datasets. TabNet combines the advantages of tree-based methods and DNN-based methods, resulting in both high performance and interpretability. By replacing DNNs with tree-based methods, the interpretability of DNNs' superior performance can be further enhanced. Drawing inspiration from this concept, we propose the use of TabNet in the context of fatal health classification in this paper.
 
\section{Methods and Materials}
The primary objective of this study is to devise a reliable method for classifying child and maternal health based on the risk of mortality using fetal health data. In a clinical environment, achieving a TabNet model accuracy of at least 94.36 \% is essential to ensure the model's effectiveness.

\subsection{Overview of the Dataset}
Table (1) presents the Cardiotocogram (CTG) features description. This dataset (ref.) that was used in this study contains 2126 records of attributes that were taken from CTG exams and classified by experts into three categories: normal, suspect, and pathological. In the dataset, the Normal, Suspect, and Pathological classes are denoted by the numbers 1, 2, and 3, respectively. Fig (1) signifies the heat map showing the correlation in the Fetal health data. There are numbers inside the small box with a different color intensity that represents pearson's correlation coefficient.
\begin{table}[htbp]
  \centering
  \caption{Cardiotocogram Features Description}
    \resizebox{8.5 cm}{!}{\begin{tabular}{ll}
    \hline
    \hline
    \multicolumn{1}{c}{Features} & \multicolumn{1}{c}{Description} \\
    \hline
    Baseline Value & FHR beats per minute \\
    Accelerations & Number of accelerations per second \\
    Fetal movement & Number of fetal movement \\
    Uterine contractions & Number of uterine contractions per second \\
    Light decelerations & Number of light decelerations per second \\
    Severe decelerations & Number of severe decelerations per second \\
    Prolongued decelerations & Number of prolonged decelerations per second \\
    Abnormal short-term variability & Percentage of time with abnormal short-term variability \\
     Short-term variability (Mean value) & Mean value of short-term variability \\
     Abnormal long-term variability & Percentage of time with abnormal long-term variability \\
    Long-term variability (Mean value) & Mean value of long-term variability \\
    Histogram width & Width of FHR histogram \\
    Histogram min & Minimum of FHR histogram \\
    Histogram max & Maximum of FHR histogram \\
    Histogram number of peaks &  Number of histogram peaks \\
    Histogram number of zeroes & Number of histogram zeroes \\
    Histogram mode & Mode value of histogram \\
    Histogram mean & Mean value of histogram \\
    Histogram median & Median value of histogram \\
    Histogram Variance & Variance of histogram \\
    Histogram tendency & Tendency of histogram \\
    Fetal Health & Fetal state (code: N=Normal, S=Suspected, P=Pathological) \\
    \hline
    \hline
    \end{tabular}}
\end{table}
\begin{figure*}
  
    \centering
    \includegraphics[width=0.83\paperwidth]{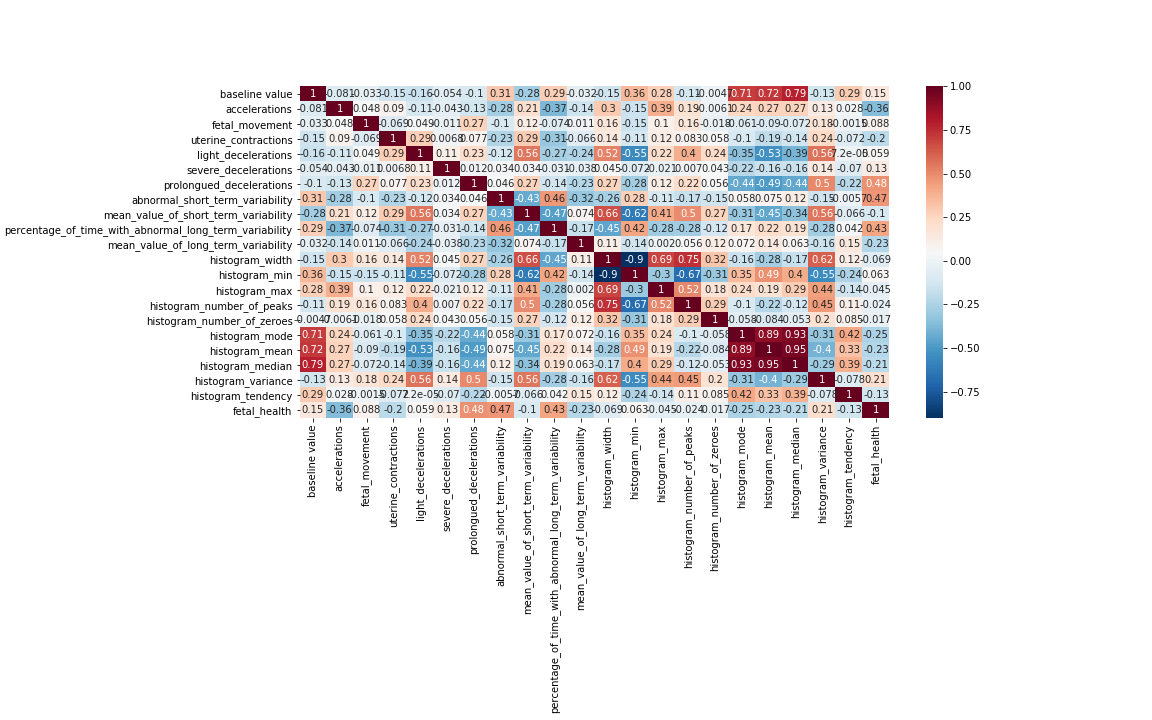}
    \caption{Heat map to show correlations in fetal health data. The numbers inside the small box represent pearson correlation coefficients.}
\end{figure*}
\section{Machine Learning Models}
\subsection{Support Vector Machine (SVM)}\label{AA}
Support Vector Machines (SVMs) were first introduced by Vladimir Vapnik and Alexey Chervonenkis in 1963. Their original work laid the theoretical groundwork for SVMs, which focused on the concept of the "margin" for binary classification. But their practical applications were significantly improved by Vapnik and Cortes in 1995 \cite{10.1023/A:1022627411411}. This method has various pattern recognition tasks, such as text recognition, object recognition, sound recognition, and face recognition. Their versatility in handling different types of data and the ability to model complex decision boundaries make them popular in many real-world applications \cite{10.1145/973264.973281}.
\subsection{Principal Component Analysis (PCA)}\label{AA1}
Principal Component Analysis (PCA) is indeed a powerful tool for feature extraction and dimensionality reduction in data analysis and machine learning. The method was first devised by J. Ross Quinlan in 1986 and later popularized by Matthew Turk and Alex Pentland \cite{139758}. The key advantage of PCA is its ability to reduce the dimensionality of the data while retaining most of the important information. It does so by arranging the principal components in order of importance, with the first component capturing the highest variance, the second component capturing the second-highest variance, and so on. Typically, a significant portion of the data's variance can be explained using just a few of the top principal components, allowing us to represent the data in a lower-dimensional space without losing much relevant information \cite{10.1111/1467-9868.00196}.
\subsection{Linear Discriminant Analysis (LDA)}\label{AA2}
Ronald A. Fisher in 1936 introduced LDA as a statistical method for dimensionality reduction and classification. He formulated the problem of finding linear combinations of variables that best discriminate between different classes in the data. Fisher's work laid the theoretical foundation for what later became known as Linear Discriminant Analysis \cite{fisher36lda}. This classification technique is used for pattern recognition and machine learning tasks. The goal of LDA is to find a linear combination of features that maximizes the separation between classes while minimizing the scatter within each class. It is widely used in various applications, including face recognition, document classification, and bioinformatics \cite{10.1145/1553374.1553460}. 
\subsection{Random Forest (RF)}\label{AA3}
Random Forests (Leo Breiman, 2001) is an ensemble learning method combining multiple decision trees. Each tree is trained on a random subset of data with replacement and considers only a random subset of features at each node. The final prediction is obtained by aggregating the outputs of individual trees through majority voting (for classification) or averaging (for regression). This approach improves model accuracy and generalization by reducing overfitting and introducing diversity among the trees\cite{breiman2001random}.
\subsection{Attentive Interpretable Tabular Learning (TabNet)}\label{AA4}
TabNet (Arik and Pfister, 2019) is a deep learning architecture specially designed for tabular data that relies on a tree-like structure, allowing for the linear combination of features by computing coefficients that determine how each feature contributes to the decision-making process \cite{shah22tab,arik20tab}. The architecture of TabNet can be represented into several key components which are shown in Figure (2). TabNet employs sparse instance-specific feature selection, which is learned during the training phase. It also constructs a sequential multi-step architecture where each decision step determines a portion of the final decision by leveraging the selected features. Additionally, it incorporates non-linear processing of features. This method is more efficient for implementation in contrast to traditional deep neural network (DNN)-based methods, as TabNet offers a robust soft feature selection ability and provides control over sparsity through sequential attention mechanisms. 
\begin{figure}
  \centering
    \includegraphics[width=0.33\paperwidth]{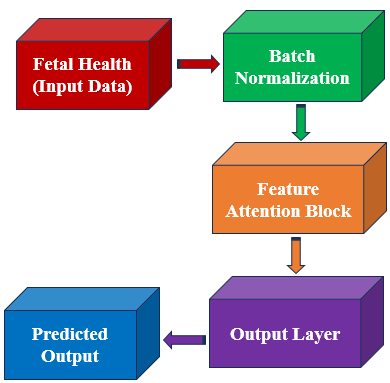}
    \caption{Block diagram of TabNet for Fetal health classification}
\end{figure}

\section{Experimental Analysis}
Following the tuning process, the suggested models demonstrated exceptional performance on the test data. The TabNet achieved an accuracy of 94.36 \%, the support vector machine with LDA achieved an accuracy of 90.42 \%, and the random forest with LDA achieved an accuracy of 91.13 \%. These highly accurate and efficient models can be deployed globally, particularly in settings where assessing fetal health individually for each case is impractical for obstetricians. Table (II) presents the classification accuracy of SVM and RF. The dimensionality reduction methods PCA and LDA were implemented on both SVM and RF classifiers. Moreover, the training size was varied in the several percentage ranges (40, 50, 60, 70, 80). For SVM, LDA shows better accuracy than PCA with 89.41\% for 80\% training size. Similarly, For RF, LDA shows optimum accuracy than PCA with 88.94\% for 80\% training size. Table (III) displays the classification results achieved by the more efficient TabNet model on Fetal health tabular data. The training dataset sizes were varied similarly to those shown in Table (II), and the accuracy percentages were predicted. Notably, TabNet outperformed SVM and RF models with PCA and LDA in terms of accuracy at each training size. This observation highlights TabNet's superior classification performance compared to RF and SVM models.

Figure (3) illustrates a histogram displaying the classification accuracies of SVM and RF with PCA and LDA, alongside the TabNet model. The data presented in this plot are from the results outlined in Table (II) and Table (III). This histogram provides a visual representation of the classification performance, confirming the superior accuracy of TabNet over SVM and RF models with PCA and LDA, as indicated in the tables. Figure (4) displays the confusion matrix generated for Fetal health data using the TabNet model trained on an 80\% subset of the dataset. The matrix clearly indicates that the TabNet model excelled in accurately predicting the 'normal' class, which signifies its proficiency in identifying typical or healthy cases within the Fetal health data.

\begin{table}[htbp]
  \caption{Classification accuracy percentage with different train sizes}
 \resizebox{\hsize}{!}{
 {\begin{tabular}
 {crccccc|ccccc}
    \hline
    \hline
    &   &    &    & PCA  &  &  &   &    & LDA  &       &  \\
    \hline
     & \multicolumn{1}{l}{Train size (\%)} & 40    & 50    & 60    & 70    & 80    & 40    & 50    & 60    & 70    & 80 \\
    \hline 
    SVM   &  & 79.19 & 79.83 & 79.86 & 81.95 & 82.82 & 90.42 & 90.10 & 89.99 & 89.17 & 89.41 \\[+0.1em]
    RF  &  & 83.75 & 84.35 & 83.62 & 83.00 & 83.05 & 91.13 & 89.91 & 90.22 & 90.26 & 88.94 \\[+0.1em]
    \hline
    \hline
    \end{tabular}}
  }
\end{table}
\begin{table}
\caption{Accuracy percentage using TabNet with different train sizes}
\begin{tabular}{cccccccc}
\hline 
\hline
& \multicolumn{1}{l}{Train size (\%)} & 40 & 50 & 60 & 70 & 80 \\
\hline TabNet & & 92.07 & 92.08 & 93.16 & 93.4 & 94.36 \\
\hline
\hline
\end{tabular}
\end{table}
\begin{figure}
   \centering
    \includegraphics[width=0.4\textwidth]{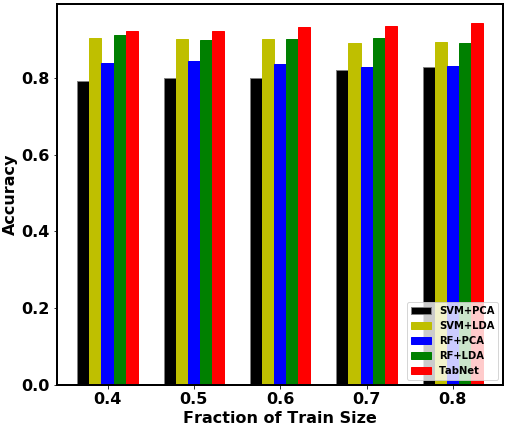}
    \caption{Change of accuracy vs train size with different algorithms}  
\end{figure}

\begin{figure}
  
    \centering
    \includegraphics[width=0.41\paperwidth]{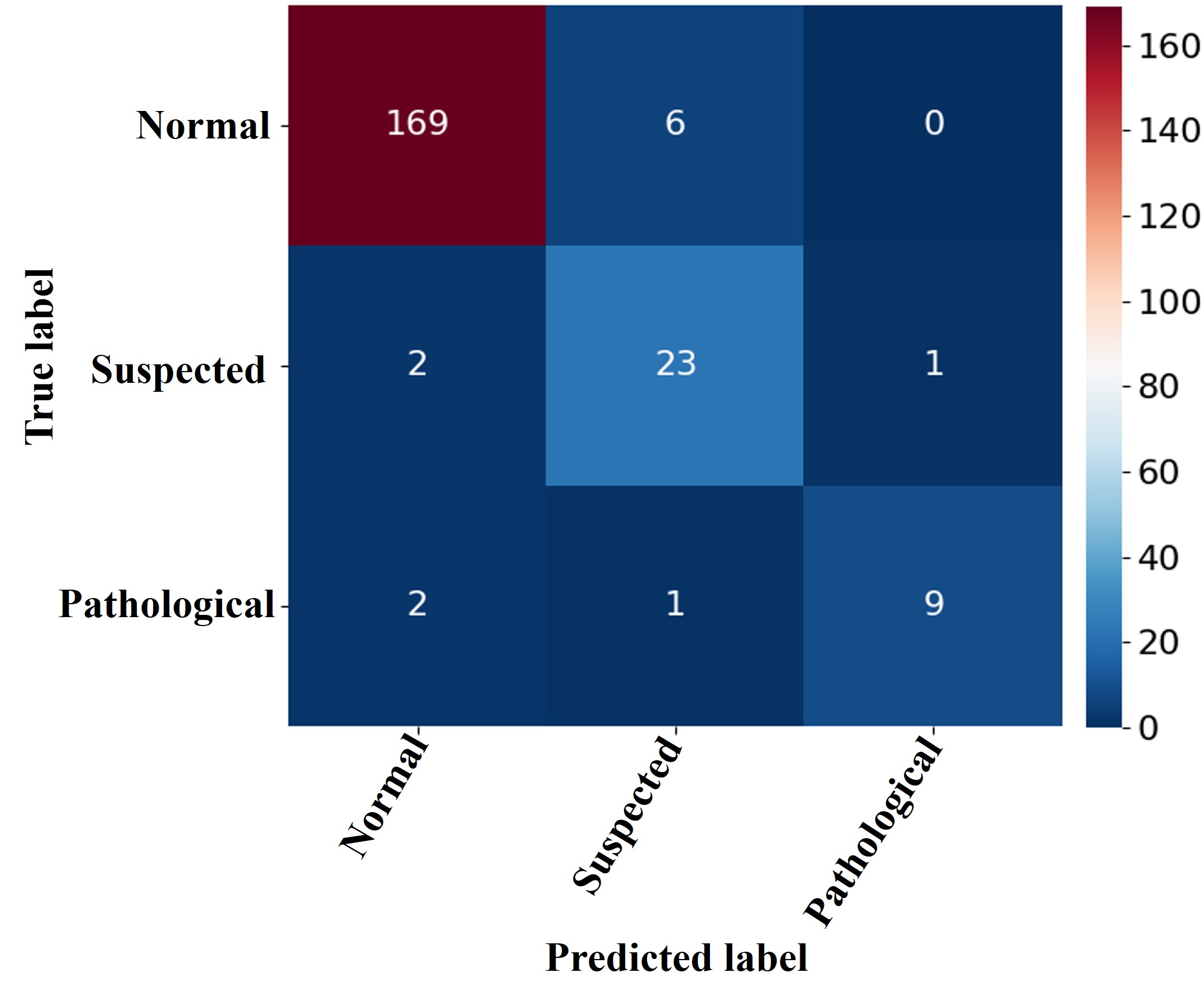}
    \caption{Confusion matrix for TabNet at 80 \% training samples on fetal health data}
   
\end{figure}

\section{Conclusion}
This study examines the influence of LDA and PCA dimensionality reduction techniques on machine learning classifiers for identifying fetal health abnormalities using CTG exams. Our results clearly illustrate that LDA performs better than PCA, and the Random Forest classifier when combined with LDA, yielded the highest performance compared to combined with PCA algorithm. It achieved an optimum accuracy of 91.13 \% in classifying prenatal health abnormalities. In addition, we also applied a deep learning method called TabNet which utilizes attention mechanisms to focus on relevant features and employs promising performance in finding the health status of the fetus. By utilizing this method, we obtained a higher accuracy rate compared to other dimensionality reduction algorithms we used. This outcome holds promising implications for cost-effectiveness, aiming to reduce maternal and fetal mortality rates. Furthermore, based on the results of this study, it can be inferred that cardiotocogram readings have the potential to predict fetal health outcomes.
\bibliography{references}

\begin{thebibliography}{10}

\bibitem{cho22fet}
A.~Chowdhury, A.~Chahar, R.~Eswara, M.~A. Raheem, S.~Ehetesham, and B.~K.
  Thulasidoss, ``Fetal health prediction using neural networks,'' in {\em 2022
  8th International Conference on Advanced Computing and Communication Systems
  (ICACCS)}, vol.~1, pp.~256--260, 2022.

\bibitem{yin23us}
Y.~Yin and Y.~Bingi, ``Using machine learning to classify human fetal health
  and analyze feature importance,'' {\em BioMedInformatics}, vol.~3, no.~2,
  pp.~280--298, 2023.

\bibitem{gill2023abnormal}
P.~Gill, J.~M. Henning, K.~Carlson, and J.~W. Van~Hook, ``Abnormal labor,'' in
  {\em StatPearls [Internet]}, StatPearls Publishing, 2023.

\bibitem{shah21col}
C.~Shah and Q.~Du, ``Collaborative and low-rank graph for discriminant analysis
  of hyperspectral imagery,'' {\em IEEE Journal of Selected Topics in Applied
  Earth Observations and Remote Sensing}, vol.~14, pp.~5248--5259, 2021.

\bibitem{gao15svm}
L.~Gao, J.~Li, M.~Khodadadzadeh, A.~Plaza, B.~Zhang, Z.~He, and H.~Yan,
  ``Subspace-based support vector machines for hyperspectral image
  classification,'' {\em IEEE Geoscience and Remote Sensing Letters}, vol.~12,
  no.~2, pp.~349--353, 2015.

\bibitem{shah22spat}
C.~Shah and Q.~Du, ``Spatial-aware collaboration–competition preserving graph
  embedding for hyperspectral image classification,'' {\em IEEE Geoscience and
  Remote Sensing Letters}, vol.~19, pp.~1--5, 2022.

\bibitem{shah22tab}
C.~Shah, Q.~Du, and Y.~Xu, ``Enhanced tabnet: Attentive interpretable tabular
  learning for hyperspectral image classification,'' {\em Remote Sensing},
  vol.~14, no.~3, 2022.

\bibitem{tin98rand}
T.~K. Ho, ``The random subspace method for constructing decision forests,''
  {\em IEEE Transactions on Pattern Analysis and Machine Intelligence},
  vol.~20, no.~8, pp.~832--844, 1998.

\bibitem{arik20tab}
S.~O. Arik and T.~Pfister, ``Tabnet: Attentive interpretable tabular
  learning,'' 2020.

\bibitem{chris06patt}
C.~M. Bishop and N.~M. Nasrabadi, ``Pattern recognition and machine learning,''
  {\em J. Electronic Imaging}, vol.~16, p.~049901, 2006.

\bibitem{li14dec}
W.~Li, S.~Prasad, and J.~E. Fowler, ``Decision fusion in kernel-induced spaces
  for hyperspectral image classification,'' {\em IEEE Transactions on
  Geoscience and Remote Sensing}, vol.~52, no.~6, pp.~3399--3411, 2014.

\bibitem{10.1023/A:1022627411411}
C.~Cortes and V.~Vapnik, ``Support-vector networks,'' {\em Mach. Learn.},
  vol.~20, p.~273–297, sep 1995.

\bibitem{10.1145/973264.973281}
M.~Grimaldi, P.~Cunningham, and A.~Kokaram, ``A wavelet packet representation
  of audio signals for music genre classification using different ensemble and
  feature selection techniques,'' in {\em Proceedings of the 5th ACM SIGMM
  International Workshop on Multimedia Information Retrieval}, (New York, NY,
  USA), p.~102–108, Association for Computing Machinery, 2003.

\bibitem{139758}
M.~Turk and A.~Pentland, ``Face recognition using eigenfaces,'' in {\em
  Proceedings. 1991 IEEE Computer Society Conference on Computer Vision and
  Pattern Recognition}, pp.~586--591, 1991.

\bibitem{10.1111/1467-9868.00196}
M.~E. Tipping and C.~M. Bishop, ``{Probabilistic Principal Component
  Analysis},'' {\em Journal of the Royal Statistical Society Series B:
  Statistical Methodology}, vol.~61, pp.~611--622, 01 2002.

\bibitem{fisher36lda}
R.~A. Fisher, ``The use of multiple measurements in taxonomic problems,'' {\em
  Annals of Eugenics}, vol.~7, no.~7, pp.~179--188, 1936.

\bibitem{10.1145/1553374.1553460}
Y.~Liu, A.~Niculescu-Mizil, and W.~Gryc, ``Topic-link lda: Joint models of
  topic and author community,'' in {\em Proceedings of the 26th Annual
  International Conference on Machine Learning}, (New York, NY, USA),
  p.~665–672, Association for Computing Machinery, 2009.

\bibitem{breiman2001random}
L.~Breiman, ``Random forests,'' {\em Machine Learning}, vol.~45, no.~1,
  pp.~5--32, 2001.

\end{thebibliography}
\bibliographystyle{ieeetr}

\end{document}